\documentclass{article}

\usepackage{arxiv}

\usepackage[utf8]{inputenc} 
\usepackage[T1]{fontenc}    
\usepackage{hyperref}       
\usepackage{url}            
\usepackage{booktabs}       
\usepackage{amsfonts}       
\usepackage{nicefrac}       
\usepackage{microtype}      
\usepackage{lipsum}

\usepackage{times}
\usepackage{latexsym}

\usepackage{times}
\usepackage{url}
\usepackage{latexsym}

\usepackage{algorithm}
\usepackage{pifont}
\usepackage{tikz}
\usepackage{amsmath}
\usepackage{pgfplots}
\usepackage{balance}
\usepackage{multibib}

\usepackage{graphicx}
\usepackage{amssymb,amsmath,bm}
\usepackage{textcomp}
\usepackage{balance}

\usepackage{subfig}

\usepackage{mathtools}
\usepackage{booktabs}
\usepackage{array,dcolumn}

\usepackage{algorithm,algpseudocode}
\usepackage{pgfplots}
\pgfplotsset{width=8cm,compat=1.8}
\usetikzlibrary{patterns}

\usetikzlibrary{shadows,arrows}
\pgfdeclarelayer{background}
\pgfdeclarelayer{foreground}
\pgfsetlayers{background,main,foreground}

\tikzstyle{materia}=[draw, fill=blue!20, text width=6.0em, text centered,
  minimum height=1.5em,drop shadow]
\tikzstyle{practica} = [materia, text width=8em, minimum width=10em,
  minimum height=3em, rounded corners, drop shadow]
\tikzstyle{texto} = [above, text width=6em, text centered]
\tikzstyle{linepart} = [draw, thick, color=black!50, -latex', dashed]
\tikzstyle{line} = [draw, thick, color=black!50, -latex']
\tikzstyle{ur}=[draw, text centered, minimum height=0.01em]
 
\newcommand{\practica}[2]{node (p#1) [practica]
  {Step #1\\{\scriptsize\textit{#2}}}}

\newcommand{\background}[5]{%
  \begin{pgfonlayer}{background}
    \path (#1.west |- #2.north)+(-0.5,0.5) node (a1) {};
    \path (#3.east |- #4.south)+(+0.5,-0.25) node (a2) {};
    \path[fill=yellow!20,rounded corners, draw=black!50, dashed]
      (a1) rectangle (a2);
    \path (a1.east |- a1.south)+(0.9,-0.28) node (u1)[texto]
      {\scriptsize\textit{#5}};
  \end{pgfonlayer}}

\title{A Hybrid Framework for Topic Structure using Laughter Occurrences}

\author{
Sucheta Ghosh\\
        HITS gGmbH,\\
        Heidelberg,\\
        69117 Germany.
}

\begin{document}
\maketitle

\begin{abstract}
Conversational discourse coherence depends on both linguistic and paralinguistic phenomena. In this work we combine both paralinguistic and linguistic knowledge into a hybrid framework through a multi-level hierarchy. Thus it outputs the discourse-level topic structures. The laughter occurrences are used as paralinguistic information from the multiparty meeting transcripts of ICSI database. A clustering-based algorithm is proposed that chose the best topic-segment cluster from two independent, optimized clusters, namely, hierarchical agglomerative clustering and $K$-medoids. Then it is iteratively hybridized with an existing lexical cohesion based Bayesian topic segmentation framework. The hybrid approach improves the performance of both of the stand-alone approaches. This leads to the brief study of interactions between topic structures with discourse relational structure. This training-free topic structuring approach can be applicable to online understanding of spoken dialogs.
\end{abstract}


\section{Introduction}
Topic structures are basically a type of discourse structure. There exists many methods for topic segmentation those use semantic, lexical and referential similarity or, more recently, the language models \cite{hearst-97,elhadad-99,utiyama-01,galley-03,malioutov-06,eisenstein-08}. An automated topic segmentation tool splits a discourse into a linear sequence of topics such as the geography of a country, followed by its history, demographics, economy, legal structures, etc.; this segmentation is usually done on a sentence-by-sentence basis, with segments not assumed to overlap \cite{webber-12}. 
In this work the whole process utilizes a stage-by-stage procedure like it is devised by \cite{grosz-86}. Thereby we achieve a discourse level topic structure, beyond the sentence level structure. In our case, segmentations with overlaps do not have much relevance because our primary level segmentation is done on the basis of shared laughter occurrences in meeting dialog; during those laughter turns there exists out-of-topic and fragmented comments. So we acquire a high level hybrid structure of topic, that combines the local and global levels of segmentation through a multi-level hierarchy of linear segmentations. 

In this work, we propose a training-free framework for conversation segmentation incorporating the psychological evidences of coherence in conversation segmentation. According to the literatures of psychology, shared vocalized laughter in a conversation invariably denotes the change of topic, the vice versa is not necessarily true, that is, in a conversation sometimes the topic changes occur without any occurrence of laughter \cite{holt-10}. Therefore, if we are able to locate the shared laughter occurrences in a conversation, it will veritably indicate the segments with same topics. 
To achieve this, we acquire the laughter occurrences from the annotations of database. Then we cluster the laughter occurrences. Based on the number of cluster membership the shared laughter occurrences are identified. The shared laughter occurrences determine the segment boundaries. To ensure the robustness of optimization two clustering techniques are applied independently: agglomerative hierarchical clustering and $K$-medoids. These techniques were optimized and integrated to achieve the best performance, namely, fixed boundary clumping for hierarchical clustering and iterative optimization for $K$-medoids. 
Then we hybridize this basic (clustering based) segmentation with a lexical cohesion based Bayesian text segmentation method \cite{eisenstein-08}. The resultant hybrid method performs better than the stand-alone approaches. The approach is fit for online setup.

The goal of this work is not merely to improve the state of the art performance combining multiple methods. This is to show that the linguistic and paralinguistic information are both useful to achieve a conversational topic discourse. Also, we conduct a brief study of interaction between topic structure and the discourse relational structure. 

\subsection{Related Works}\label{sec:relwrk}

A significant number of segmentation techniques are based on lexical cohesion \cite{hearst-97,elhadad-99,utiyama-01,galley-03,malioutov-06,eisenstein-08}. 
\cite{halliday-76}'s seminal work states about two main types of cohesion, namely (1) grammatical, which is based on structural content, and (2) lexical, which is based on lexical content and background knowledge. There exists some grammatical cohesion based segmentation techniques \cite{giora1983,bestgen2000,taboada-09}. 
While most of the prominent works in the conversational segmentation are based on lexical cohesion, the cue word or phrase based segmentation techniques are also prevalent. The theory by \cite{grosz-77} proposes the usage of some cue word or phrase as discourse markers. 

There exists some linear conversation segmentation techniques \cite{hearst-97,kaufman1999,eisenstein-08}, but there also exists some works on the hierarchical segmentation \cite{eisenstein-09,simon2015}. In both the cases the methods based on lexical cohesion and Bayesian learning are most successful.

The task of discourse segmentation started in a supervised setup with written texts using cue words or cue phrase discourse marker \cite{grosz-77,litman-93,elsner-08}. The problem is that these studies are constrained with respect to the important factors like available resources, dealing with the problems of spoken language wide variability, dealing with the multi-lingual setup.
%

\cite{hearst-97} defined subtopic groupings for science-news article paragraphs into some classes, which evolved into present day templates that can be filled with proper values. This work called TextTiling, is basically a cohesion based approach. The work of \cite{choi-00} used a divisive clustering algorithm to linearly segment the text. In the unsupervised model of \cite{galley-03}, inference is performed by selecting segmentation points at the local maxima of the cohesion function. \cite{malioutov-06} optimized a normalized minimum-cut criteria based on a variation of the cosine similarity between sentences.

There also exists a series of notable works those use Bayesian approach to linearly segment the topic. As we know that Bayesian inference is fit for dynamic analysis of sequential data, and also it is easier to incorporate additional features or resources like language model with a Bayesian framework. The work of \cite{eisenstein-08} showed an way of unsupervised topic segmentation using Bayesian approach , whereas the work of  \cite{utiyama-01} was a special case of Bayesian learning; an alternative Bayesian approach was also proposed by \cite{purver-06}.

Among all the other existing approaches the lexical cohesion based unsupervised Bayesian approach by \cite{eisenstein-08} is one of the important frameworks. In their work the authors show that the lexical cohesion can be placed in a Bayesian context by modeling the words in each topic segment as it draws from a multinomial language model associated with the segment; maximizing the observation likelihood in such a model it yields a lexically-cohesive automatic segmentation \cite{eisenstein-08}. The Bayesian framework provides a principled way to incorporate additional features such as cue phrases, which is a powerful indicator of discourse structure in the unsupervised segmentation frameworks.  
This model still offers consistent improvements over an array of state-of-the-art systems on both text and speech datasets. Therefore we choose the work as one of our baselines. 


We have described the topic segmentation methods for written texts, which is either a well structured document or a semi-structured or unstructured transcribed conversation. There also exist seminal works on spoken discourses: both of \cite{hirschberg-96} and \cite{niekrasz-09} use the prosodic phrases, the latter used linguistic features and the former used low level paralinguistic, prosodic features like pitch, duration. The accurate extraction of low-level prosodic features is still a challenging area. 


In this work we also aim at incorporating the knowledge from cognitive psychology, by the means of human laugh occurrences. Prior works of \cite{holt-10}, \cite{bonin-12} already stated that there is an evident link between the topic change and laughter in human-human discourses. Studies also report that the voiced laughter generates positive affect \cite{bachorowski-01}; the mirthful laughter activities have been suggested as modifiers of neuroendocrine hormones involved in the classical stress response \cite{berk-89}, and the stress has a negative effect on human attention \cite{jensen-98}. In the meetings, people generally use laughter as a de-stressing activity and then move on to a new discourse or topic. It may also be possible that a meeting does not contain any laughter situation, this study does not cover that situation, it only covers the natural human conversations with laughter.

\subsection{Motivation: hybrid segmentation of conversation using laughter}\label{sec:mot}
Spoken language comprehension does not depend only on the text or the words. The cognitive model of language comprehension by \cite{bower-85} states about a step-by-step comprehension process. On the first steps the words are recognized; on the subsequent steps the prosody and gestures are involved, thus finally the whole matters are comprehended through the extraction of features and pattern recognition. There exists many paralinguistic features those are actively involved in this process of comprehension.  Among all paralinguistic features, laughter surely plays an important role, that we already discussed in \ref{sec:relwrk} citing the several works, and also there we mentioned about our preference for the stage by stage processing for various kinds of informations. 

It is interesting to conduct a study of interaction of the topic structure and any other discourse structures, namely discourse relational structure. On the other hand, the real-time classification in spoken language understanding seeks those approaches, which can also be used as online framework. In this case, we do not wait until all the data is captured before starting the online process of understanding. The frameworks internally process data, segment by segment with no dependency on the future. Thus if we can acquire the transcripts through an online Automatic Speech Recognizer \cite{kaldionline16} and also if we be able to achieve laugh occurrences\cite{salamin13,schuller13,ghosh16}, then we are able to acquire this topic structure online. 


\subsection{Our Contribution}\label{sec:cont} 
In this work, we present a framework that follows a cascaded process that hybridizes two standalone processes in a hierarchy: at the first level we segment a conversation using paralinguistic information laughter, then at the next level, we include a third party tool for topic segmentation of the text, subjected to the primary paralinguistic segmentation. We observe a significant improvement of result of this hybrid framework in comparison to the standalone techniques. The goal of this hybridization is not merely to improve the baseline; through this work we propose a framework that not only works along with the cognitive model of human comprehension \cite{bower-85}, but also it improves the baseline. Thus there remains the possibility to use the same framework in different context of spoken language comprehension. Additionally, here we study on the interaction between the topic and discourse relational structures.

\section{Laughter based Hybrid Segmentation of Conversation}\label{sec:hybseg}
\begin{figure}
\begin{tikzpicture}[scale=0.80,transform shape]
 
%
  \path \practica {Input}{INPUT: array of turn-indices};
  \path (pInput.south)+(-2.5,-1.25) \practica{1a}{Agglomerative Hierarchical Clustering};
  \path (pInput.south)+(2.5,-1.25) \practica{1b}{$K$-medoids Clustering};

  \path (p1a.south)+(0.0,-1.3) \practica{2a}{Fixed Boundary Clumping};
  \path (p1b.south)+(0.0,-1.5) \practica{2b}{Iterative Process};
  \path (p2b.south)+(-2.5,-1.5) \practica{3}{BestCluster: Best performing Cluster Segmentation};
 \path (p3.south)+(0.0,-1.5) \practica{4}{Hybrid: Another Iterative run of BayesSeg on segments of BestCluster};
 \path (p4.south)+(0.0,-1.5) \practica{Output}{Array of clustered turn-indices}; 
     

     

  \path [line] (pInput.south) -- +(0.0,-0.25) -- +(-2.5,-0.25)
    -- node [above, midway] {} (p1a);
  \path [line] (pInput.south) -- +(0.0,-0.25) -- +(+2.5,-0.25)
    -- node [above, midway] {} (p1b);

  \path [line] (p1a.south) -- node [above] {} (p2a) ;
  \path [line] (p1b.south) -- node [above] {} (p2b) ;


  \path [line] (p2b.south) -- +(0.0,-0.25) -- +(-2.42,-0.25)
    -- node [above, midway] {} (p3);
  \path [line] (p2a.south) -- +(0.0,-0.3) -- +(+2.5,-0.3)
    -- node [above, midway] {} (p3);
    
  \path [line] (p3.south) -- +(0.0,-0.5) -- node [above] {} (p4) ;     
  \path [line] (p4.south) -- node [above] {} (pOutput) ;
   
  \background{p1a}{p1b}{p1b}{p1a}{Cluster}
  \background{p1a}{p2a}{p1b}{p2b}{Optimize}
  \background{p1a}{p3}{p1b}{p3}{BestCluster}
 \background{p1a}{p4}{p1b}{p4}{Hybridize}
\end{tikzpicture}
\caption{\footnotesize The diagram of basic processes of the whole hybrid framework}\label{fig:proc}
 \end{figure}
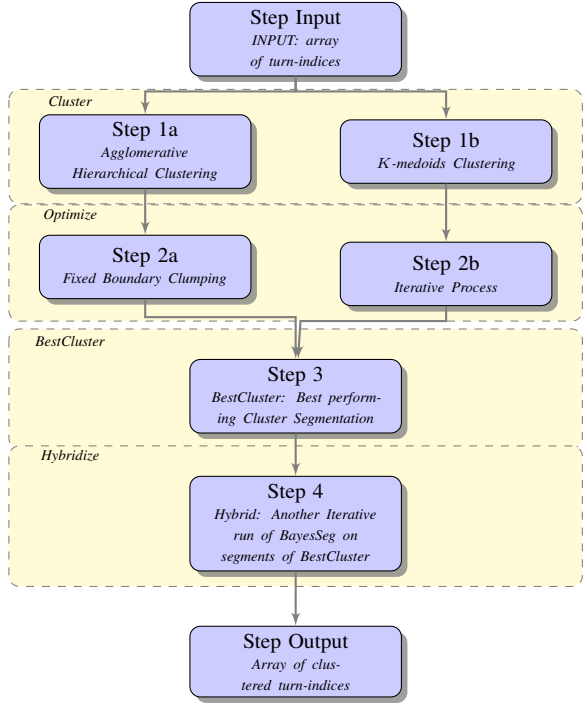

The two simultaneous way of processes and a four-step approach is followed to achieve this hybrid segmentation of the conversational discourse. We depict all four stages and two simultaneous processes in the Figure \ref{fig:proc}.

First, an initial segmentation is achieved using agglomerative hierarchical clustering technique; we use clustering because it is a flexible technique, in worst case this is better than any chance experiment, specifically the hierarchical clustering produces deterministic results and stores much information in its structure.

We use two kinds of clustering to make the decision robust. In parallel to the hierarchical clustering, we employ another segmentation technique based on $K$-medoids clustering using the same input of the agglomerative clustering technique. $K$-medoids clustering is a flat clustering technique like $K$-means clustering. $K$-means is a classical robust partitioning technique of clustering that clusters the data set of n objects into $K$ clusters if $K$ known \emph{a priori}. The $K$-medoids algorithm is a clustering algorithm related to the $K$-means algorithm and the median-shift algorithm \cite{shapira-09}. Median-shift is a mode seeking algorithm that relies on computing the median of local neighborhoods, instead of the mean.

In contrast to the $K$-means algorithm, $K$-medoids chooses datapoints as centers, and in case of $K$-means the distances between the centroids of cluster set are calculated using $L2$-norms whereas in case of the $K$-medoids it uses $L1$-norms to calculate the distance between the medoids. The details of the method is given in the next section \ref{sec:kmclus}. Likewise the $K$-means, The $K$-medoids algorithm algorithm uses an iterative refinement technique for the optimization of performances.
As a second step, we optimize the primary segmentation of the agglomerative clustering through a fixed boundary clumping following \cite{niekrasz-10}, and as well, we optimize the primary segmentation of the K-medoids clustering through an iterative process.
In the third step, we choose the best performing segmentation between the agglomerative and $K$-medoids clustering. 
Finally, in the fourth step we hybridize this clustering based framework with the lexical cohesion based Bayesian text segmentation framework. Then we obtain the output of segmented conversation by means of the clustered turn-indices.  

\vspace{-0.01em}
 \vspace{-0.01em}

\subsection{Segmentation using Agglomerative Clustering}\label{sec:hclus}
This work uses multiparty meeting conversations. Each document is defined as a linear sequence of turns occurring from many participants. Each document is a global discourse, whereas we aim at segmenting this whole document into linear chunks, that is for a M-length sequence of turns $\langle t_1, t_2 \cdots t_M \rangle$  in a global discourse, minimal local discourse boundaries are $\langle [0,t_1\},\{t_1,t_2\}, \cdots \{t_{M-1}t_M\}, \{t_M, D]\rangle$, $0$: the starting point, $D$: ending point. A potential segmentation $X$ can be defined as a collection of non-overlapping subsequences of the M-length sequence. Thus if we collect, say N-length turns with laughter occurrences, then we may primarily segment the $M$-length turn sequence into $N+1$ segments.
%

Between each pair of laughter-turn indices $\{l_i, l_j\}$ a distance measure is calculated, thus we collect an $N\times N$ distance matrix. We use the euclidean distance criteria. Then linkage criterion determines the distance between the sets of observations as a function of the pairwise distances between observations. We use the average linkage criteria. Thus we acquire a hierarchy tree $Z$, that contains paired objects and their links. 

To determine the exact number of clusters, specifically, to determine the cutoff of the hierarchy tree, we compute inconsistency coefficient for each link of the cluster tree. The relative consistency of each link in a hierarchical cluster tree can be quantified and expressed as the inconsistency coefficient. Links that join distinct clusters have a high inconsistency coefficient; links that join indistinct clusters have a low inconsistency coefficient. For each link $k$, this coefficient is computed as: $$Y(k)=\frac{Z(k) - m}{std}$$ where m: the mean of the heights of all the links included in the calculation, std: the standard deviation of the heights of all the links included in the calculation. We set the floor value of the maximum inconsistency as the cutoff of the cluster \cite{jain-88}.

\subsubsection{Segment Optimization using Fixed Boundary Clumping}\label{sec:opthclus}
We aim at optimizing the primary segmentation through a fixed boundary clumping following \cite{niekrasz-10}. We use this with the intuition that when the laughter turns occur adjacently, it always denotes a shared laughter due to a single local discourse segment, therefore we need to clump the adjacent laughter turns together. We follow a simple algorithm to achieve this:
this algorithm accepts a sequence of indices of the clustered segments; it clumps the adjacently occurring turn indices to merge all adjacently occurring turn indices into the last one in that adjacent series. 
This algorithm outputs optimized segmentation boundaries of discourse. 

\subsection{Segmentation using $K$-Medoids Clustering}\label{sec:kmclus}
In the previous approach of clustering we have chosen hierarchical agglomerative clustering in the section \ref{sec:hclus}, it is primarily because of its consistent segmentation of clusters. On the other hand, it was hard to choose the way to measure the dissimilarities between groups in the form of linkages. So, we propose to use other robust clustering technique like $K$-medoids on the same vector of laughter turn indices. $K$-medoids partition the vector into $K$ clusters. For each of the partition, the sum is minimized over all the clusters, of within cluster sums of a point of the vector and the cluster medoids. 

Medoids are representative objects of a data set or a cluster with a data set whose average dissimilarity to all the objects in the cluster is minimal. Medoids are similar in concept to means or centroids, but medoids are always the members of the given vector.  

Suppose that $n$ objects having $p$ variables each should be grouped into $k (k < n)$ clusters, where $k$ is assumed to be given. Let us define $j$th variable of object $i$ as $X_{ij}$ $(i = 1,\cdots,n; j = 1,\cdots,p)$. The Euclidean distance will be used as a dissimilarity measure in this study, though other measures can also be adopted. The Euclidean distance between object $i$ and object $j$ is given by, 

$$d_{ij} = \sqrt {\sum_{a=1}^{p} (X_{ia}-X_{ja})}; i=1\cdots n; j = 1 \cdots n$$

We use the proposed method by \cite{park-09} of choosing the initial medoids. This method tends to select $K$ most middle objects as initial medoids. We optimize the $K$-medoids clustering using an iterative optimization technique.

\subsubsection{Iterative Optimization of $K$-Medoid Clustering}\label{sec:optkmclus}
We set the initial value of this iterative optimization technique from the default parameter settings of one of our baseline, namely Bayesian learning based text segmentation tool, available over Internet. The proposed iterative optimization algorithm is composed of the following steps:

\begin{itemize}
\item run the $K$-medoids algorithm using default value $K=n$, where $1<n<20$\footnote{We assume that in a $60$ minutes of a meeting there cannot exist more than 20 topics on an average. The time of commencement of each meeting session in ICSI meeting corpus is about one hour, within this period around 10 mins of time is used to read the digit transcripts. The details of the task is given in the data section \ref{sec:data}.}. and store the value of evaluation metrics $P_k$ and $WD$.
\item for $(K=n-1;n-1>1;n=n-1)$, run the $K$-medoids algorithm to store corresponding $P_k$ and $WD$
\item for $(K=n+1;n+1<20;n=n+1)$, run the $K$-medoids algorithm to store corresponding $P_k$ and $WD$
\item choose the value of K where min($P_k$ and $WD$).
\end{itemize}

As a next step we hybridize the best performing cluster framework for each conversation between the optimized agglomerative and optimized $K$-medoids clustering technique. We observe that in around $60\%$ of cases of ICSI meeting conversation the optimized $K$-medoids is the best performing one.

\subsection{Hybrid Framework for Conversation Segmentation}\label{sec:hyb}

We propose an algorithm for hybridization of the clustering and lexical cohesion based Bayesian approach. We use the best performing clusters as the basis segmentation. Then we iteratively run further the lexical cohesion based Bayesian segmentation technique over each of the segments and evaluate the $P_k$ and $WD$ metrics. We choose the segmentation corresponding to the minimized value of the $P_k$ and $WD$ metrics.

The proposed algorithm is given as follows:
\begin{itemize}
\item Initialize framework with best clustering based $s$ segments: $\{1,n_1,n_2,\cdots,n_{s-1}\}$
\item for each segment: $\{1,n_i\}$, where $i=1,\cdots,s-1$, run the lexical cohesion based Bayesian approach of text segmentation to compute the $P_k$ and $WD$ metrics.
\item if the best score is found at $i=s-1$, then stop and compute the final score, 
\item elseif the min. score is found at $i < s-1$ then, further iteratively form segments using pairwise points keeping the 1st position fixed: $(n_i,n_{i+1}); \cdots (n_i,n_{s-1})$ , run the lexical cohesion based Bayesian approach of segmentation to compute the $P_k$ and $WD$ metrics.
\item choose the minimum value of ($P_k$ and $WD$)
\end{itemize}

Thus we obtain the hybrid structure of segmentation. One example: if input data is $s$ segments: $\{1,n_1,n_2,\cdots,n_{s-1}\}$ then the output segmentations can be $S$ segments: 
$\{1,n_1,x_{11},x_{12},\cdots,x_{1N},n_{N+2},\cdots,n_{S-1}\}$, 
where $S=s+N$ and $\{x_{11},x_{12},\cdots,x_{1N}\}$ is a fine-grained local segmentation achieved through the lexical cohesion based Bayesian segmentation tool. This hybrid segmentation technique works significantly better than the stand-alone baseline segmentation techniques. 

\section{Experimental Setup}\label{sec:expt}
\subsection{Data}\label{sec:data}
We use the ICSI corpus of multi-party meeting transcripts \cite{janin-03}. 
Each meeting contains between 3 and 9 participants. For each meeting, a small XML file is stored describing some meeting-specific information such as date and time stamp, participants and their turns etc. 
Laughter is annotated as discrete events in this corpus. The discrete events are annotated as VocalSound instances, and appear interspersed among lexical items. Their location among such items is indicative of their temporal extent. Among the 10 types of laugh annotated in this corpus we consider only ``laugh'' markers for this work; it is seen that $87\%$ vocal sounds are laugh in the ICSI corpus. Here we do not consider the annotation of laugh while talking. It is seen that $8.6\%$ of time is spent on laughing and an additional $0.8\%$ is spent on laughing while talking.
%
This dataset includes transcripts of $75$ multi-party meetings, of which $19$ $Bmr$ (Meetings Recorder) series meetings are used in this work. The Meeting Recorder ($Bmr$) meetings are concerned with the ICSI Meeting Corpus. The ground truths of conversation segmentation are already existing, which are annotated and prepared by \cite{galley-03}, used for evaluation only. Beside the meeting discussion each meeting has a digits reading part in recordings, we do not use that part in this work. 
\subsection{Metrics}\label{sec:metrk}
We use the $P_k$ and WindowDiff ($WD$) measures to evaluate our system \cite{beeferman-99,pevzner-02}. The $P_k$ measure estimates the probability that a randomly chosen pair of words within a window of length $k$ words is inconsistently classified. The WD metric is a variant of the $P_k$ measure, which penalizes false positives on an equal basis with near misses. Since both of these metrics are penalties, therefore the lower values indicate better segmentations. This evaluation source code is provided by \cite{eisenstein-08}.
\subsection{Baseline}\label{sec:bs}
We compute two baselines for this task: (1) we compute the unsupervised Bayesian topic segmentation by \cite{eisenstein-08}. The source code is provided by the authors. We refer this as \emph{BayesTopic} later on. (2) We use clustering based segmentation techniques. We refer this as a \emph{BestCluster} approach. Under this scheme we use two kinds of clustering for segmentation of a conversation namely, (a) hierarchical agglomerative clustering (b) $K$-medoids clustering. We choose the best performing clustering scheme for each conversation segmentation, thus we get the best performing clustering. This scheme we follow to achieve a robust segmentation through clustering approach. We use the laugh occurrence turn indices as the input for both the stand-alone baselines, i.e. \emph{BayesTopic}, \emph{BestCluster}.
\subsection{Our Setup}\label{sec:oset}
Our proposed hybrid method takes an input of the indices of the laughter turns of a conversation as laughter sequences. It outputs the index of the turns as a segment sequence. This method does not need any parameter settings.


\subsection{Results and Discussions}
We assume that in the multiparty conversations shared laughter generally signals the change of topic \cite{holt-10}. Therefore we extract all the turns annotated with vocal sound as laugh from the data transcripts; this does not consider the breath-laugh or laugh-breath annotation, because it generally denotes the solo laughter, which does not necessarily denote topic changes. 
We present a representative scatterplot of meeting $Bmr026$ in the Figure. \ref{fig:eg}. The laughter turn indices are plotted against its sequence of the occurrences. It shows a nearly step-wise function, in all the other used data we observe this pattern. On this graph we also show the optimized cluster points with different symbols and colors: almost all the shared laughters in this meeting are clustered together, solo laughters belong to the other different clusters. In all the cases this proposed segmentation technique divides the spoken digits part from the discussion part in the ICSI corpus. 

\begin{center}
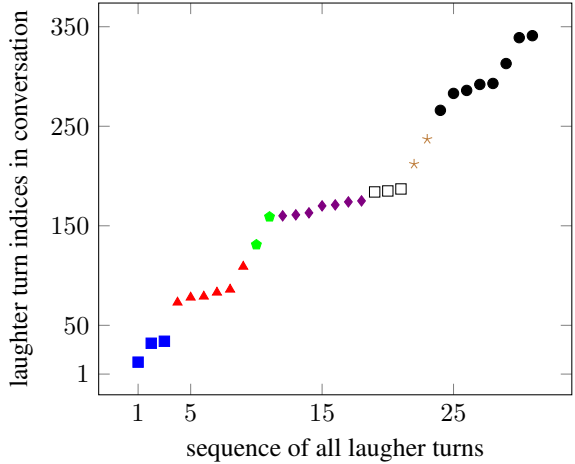
\begin{figure}
\begin{tikzpicture} [scale=.98]
\begin{axis}[
    xtick={1,5,15,25,35},
    ytick={1,50,150, 250,350},
    xlabel={sequence of all laugher turns},
    ylabel={laughter turn indices in conversation}]
\addplot[
    scatter,
    only marks,
    point meta=explicit symbolic,
    scatter/classes={
      a={mark=square*,blue},%
      b={mark=triangle*,red},%
      c={mark=pentagon*,green},%
      d={mark=diamond*,violet},%
      e={mark=square,black},%
      f={mark=star,brown},%
      g={mark=o,draw=black}},
]
table[meta=label] {
x	y	label
1	13	a
2	32	a
3	34	a
4	73	b
5	78	b
6	79	b
7	83	b
8	86	b
9	109	b
10	131	c
11	159	c
12	160	d
13	161	d
14	163	d
15	170	d
16	171	d
17	174	d
18	175	d
19	184	e
20	185	e
21	187	e
22	212	f
23	237	f
24	266	h
25	283	h
26	286	h
27	292	h
28	293	h
29	313	h
30	339	h
31	341	h
};

\end{axis}
\end{tikzpicture}
\caption{\footnotesize The scatterplot of $Bmr026$ conversation with laughter indices vs. number sequences of all laughter occurrences. The different symbols \& colors denote the shared laughter cluster (for e.g. the first cluster of shared laughter is denoted by $3$ blue cubes within first $50$ turn occurrences).}\label{fig:eg}
\end{figure}
\end{center}

We present our results in the Table \ref{tab:res}. The first row and the fourth row present the baseline performances where we used the \emph{BayesTopic} and \emph{BestCluster} algorithm. The second and third rows present our proposed clustering based approaches, \emph{OptimAggloCluster} refers to the approach of segmentation using agglomerative clustering, then optimized with fixed boundary clumping and \emph{OptimKMedoidCluster} refers to the performances of iteratively optimized of the $K$-medoids clustered segmentation. We present the results with two metrics in two columns. 
For all the five cases the $P_k$ measure performances marginally work better than the corresponding $WD$ measure performances. Between the two baselines, the \emph{BayesTopic} was harder to overcome. Between two standalone cluster-based approaches (second and third rows of Table \ref{tab:res}) \emph{OptimKMedoidCluster} performs better than the \emph{OptimAggloCluster} approach. The \emph{BestCluster}, which is our second baseline, shows better performance than the standalone cluster based approaches but it shows lower performance than the first baseline the \emph{BayesTopic}. The best performance is performed by our proposed hybrid approach \emph{hybridFrameWrk} that beats both of the baselines for both of the $P_k$ and $WD$ measures. 
It is possible to use this method without any change for a range of conversational applications like hybrid, online, multilingual and low resource language systems.

\begin{table}
\begin{center}
\small{
\begin{tabular}{l*{6}{c}r}
\hline \hline
Algorithm  & $P_k$ & $WD$  \\
\hline \hline
BayesTopic \cite{eisenstein-08} & 0.239 & 0.312 \\
\hline
OptimAggloCluster  & 0.388 & 0.404  \\
OptimKMedoidCluster  & 0.324 & 0.388 \\
BestCluster & 0.317 & 0.379 \\
\hline
hybridFrameWrk (proposed) & 0.190 & 0.248 \\
\hline
\end{tabular}
}
\end{center}
\vspace*{-\baselineskip}
\caption{The baseline and proposed approach results}\label{tab:res}
\end{table}

We observe from Table \ref{tab:indvres} that the hybrid segmentation technique also effective for the individual conversation, more than 90\% of time the performance is improved by the hybrid framework compared to the standalone performances. We notice from the same Table \ref{tab:indvres} that among all the nineteen conversations the conversations $Bmr007$ and $Bmr008$, the proposed hybrid algorithm does not perform better than the \emph{BayesTopic} algorithm, for both the metrics $P_k$ and $WD$. Still the performances are comparable to the best performances in each of these two cases. 
In the same table we also compute the means and the standard deviation of all the measures in total. We observe that comparatively the standard deviation of the window difference ($WD$) of the \emph{BayesTopic} approach is higher than that of other approaches, whereas the $P_k$ of the same approach is marginally higher than the other values of the two rest approaches.

\begin{table}[htbp]
  \centering
  \resizebox{\columnwidth}{!}{
  \small{
    \begin{tabular}{rcccccc}
    \hline
    \textbf{ICSI ID } & \multicolumn{2}{c}{\textbf{BayesTopic}} & \multicolumn{2}{c}{\textbf{BestCluster}} & \multicolumn{2}{c}{\textbf{Hybrid}} \\
   \hline
    \hline
     & P\_k & WD & P\_k & WD & P\_k & WD \\
    \hline
    Bmr001 & 0.322 & 0.431 & 0.312 & 0.421 & 0.216 & 0.325 \\
    Bmr002 & 0.243 & 0.301 & 0.351 & 0.402 & 0.244 & 0.295 \\
    Bmr005 & 0.289 & 0.374 & 0.296 & 0.381 & 0.186 & 0.271 \\
    Bmr007 & 0.159 & 0.229 & 0.398 & 0.467 & 0.174 & 0.244 \\
    Bmr008 & 0.065 & 0.076 & 0.3 & 0.31 & 0.077 & 0.088 \\
    Bmr009 & 0.299 & 0.426 & 0.325 & 0.342 & 0.247 & 0.265 \\
    Bmr010 & 0.265 & 0.292 & 0.359 & 0.387 & 0.236 & 0.266 \\
    Bmr011 & 0.269 & 0.296 & 0.373 & 0.39 & 0.261 & 0.264 \\
    Bmr012 & 0.258 & 0.475 & 0.175 & 0.379 & 0.125 & 0.329 \\
    Bmr013 & 0.247 & 0.285 & 0.161 & 0.225 & 0.085 & 0.149 \\
    Bmr014 & 0.175 & 0.253 & 0.297 & 0.393 & 0.166 & 0.244 \\
    Bmr018 & 0.326 & 0.42 & 0.297 & 0.393 & 0.21 & 0.288 \\
    Bmr020 & 0.156 & 0.239 & 0.358 & 0.477 & 0.151 & 0.216 \\
    Bmr022 & 0.025 & 0.121 & 0.331 & 0.342 & 0.112 & 0.123 \\
    Bmr024 & 0.349 & 0.458 & 0.313 & 0.407 & 0.204 & 0.258 \\
    Bmr025 & 0.253 & 0.328 & 0.28 & 0.298 & 0.266 & 0.315 \\
    Bmr026 & 0.231 & 0.286 & 0.239 & 0.294 & 0.096 & 0.151 \\
    Bmr027 & 0.295 & 0.317 & 0.436 & 0.452 & 0.292 & 0.294 \\
    Bmr029 & 0.332 & 0.332 & 0.432 & 0.432 & 0.259 & 0.336 \\
     \hline
    \textbf{Mean} & \textbf{0.239} & \textbf{0.312} & \textbf{0.317} & \textbf{0.379} & \textbf{0.19} & \textbf{0.248} \\
    \hline
    \textbf{StdDev} & \textbf{0.088} & \textbf{0.106} & \textbf{0.064} & \textbf{0.064} & \textbf{0.063} & \textbf{0.072} \\
    \hline\hline
    \end{tabular}%
    }}
 \caption{BayesTopic, BestCluster and Hybrid Approach Individual Conversation Analysis Results}\label{tab:indvres}%
\end{table}%

We run the significance testing on lexical cohesion based bayesian approach for text segmentation versus the proposed hybrid framework based on human laugh occurrences in a conversation, we find that the improvement of the performance is significant at the $95\%$ significance level.

We present the number of segments used for the best results for the individual conversations for two stand-alone and hybrid approaches in the Table \ref{tab:segres}. We perform Spearman's $\rho$ correlation test on each pairwise columns. We find that there is least ($\rho=0.10$) correlation between the \emph{BayesTopic} and \emph{BestCluster} number of segment distribution, on the other hand there is more ($\rho=0.46$) correlation between the \emph{BestCluster} and Hybrid approach the number of segment distribution, whereas the segments between the \emph{BayesTopic} and Hybrid approach are also not well-correlated ($\rho=0.26$). We also notice from the Table \ref{tab:segres} that there is no visible correlation between time of running the meeting and the number of discourse-topic segments of conversation. So in case a meeting runs much longer than the others, it does not necessarily mean that there are many local discourse or many topics to discuss. It is an interesting observation in the Table \ref{tab:segres} that for the conversation $Bmr009$ the hybrid algorithm detects only one segment for the best performance. We observe from the $Bmr009$ transcript that the participants tried to make the meeting short with two discussion-topics namely, signal processing and transcription issues, then the individual digit recording started.  

\begin{table}[h!]
\centering
\resizebox{\columnwidth}{!}{
 \begin{tabular}{||c c c c c||} 
 \hline
 ICSI ID & Time(min) & BayesTopic & BestCluster & Hybrid \\ [0.5ex] 
 \hline\hline
 Bmr001 & 36 & 7 & 2 & 4 \\ 
 Bmr002 & 46 & 7 & 4 & 4 \\
 Bmr005 & 71 & 5 & 8 & 5 \\
 Bmr007 & 60 & 4 & 8 & 2 \\
 Bmr008 & 68 & 3 & 5 & 2 \\ 
 Bmr009 & 50 & 7 & 4 & 1 \\
 Bmr010 & 54 & 5 & 6 & 9 \\
 Bmr011 & 67 & 5 & 8 & 5 \\
 Bmr012 & 42 & 6 & 7 & 5 \\
 Bmr013 & 48 & 5 & 7 & 5 \\
 Bmr014 & 50 & 4 & 4 & 4 \\
 Bmr018 & 57 & 5 & 8 & 8 \\
 Bmr021 & 37 & 6 & 9 & 5 \\
 Bmr022 & 50 & 7 & 7 & 3 \\
 Bmr024 & 52 & 9 & 12 & 6 \\
 Bmr025 & 33 & 7 & 5 & 8 \\
 Bmr026 & 45 & 5 & 4 & 4 \\
 Bmr027 & 49 & 5 & 5 & 3 \\
 Bmr029 & 43 & 3 & 2 & 3 \\ [1ex] 
 \hline
 \end{tabular}
 }
 \caption{Number of segments used for BayesTopic, BestCluster and Hybrid Approach for the Best Results (including the time for each of the meetings)}\label{tab:segres}%
\end{table}

\begin{figure}%
    \centering
    \subfloat[$P_k$ Comparison]{{\includegraphics[width=6.5cm]{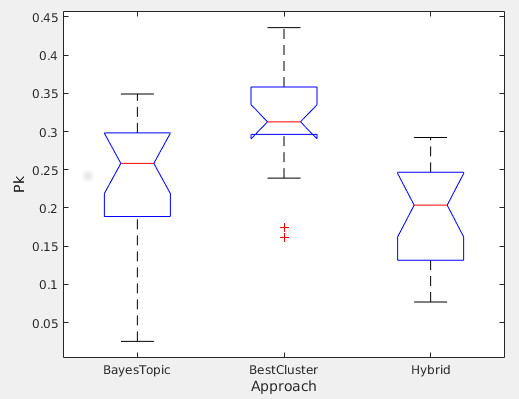} }}%
    \qquad
    \subfloat[$WD$ Comparison]{{\includegraphics[width=6.5cm]{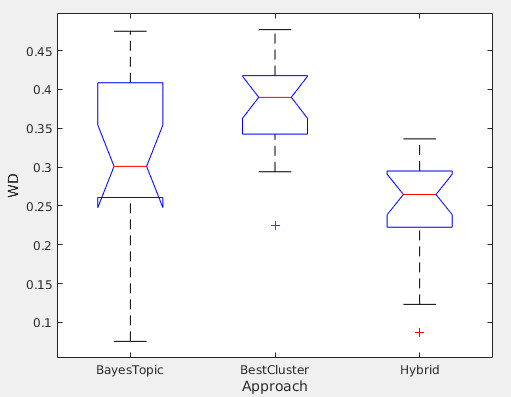} }}%
    \caption{Boxplot Comparison: BayesTopic vs. BestCluster vs. Hybrid Approaces}%
    \label{fig:3boxplot}%

\end{figure}

Both of the box-plots in the Figure \ref{fig:3boxplot} show significant improvement of the performance of the Hybrid approach over the stand-alone approaches. We observe skewness of \emph{BestCluster} performance with the $P_k$ metrics; on the other hand the \emph{BayesTopic} performance is skewed for $WD$ metrics.  The inter-quantile ranges of the \emph{BayesTopic} performance are longer than the other two approaches, for both of the metrics. We also observe the existence of outliers in the case of \emph{BestCluster} performance for both of the box-plots, whereas we find one outlier data for hybrid approach during the $WD$ box-plot comparison.

In addition, we also tested the modality of the distribution of the $P_k$ and $WD$ for set of partitions of each conversation. We plot the log-normal plot of cumulative values each of of $P_k$ and $WD$ against the time scale, we get a straight line plot from each of the used meeting conversation, which proves the uni-modality of each of the conversation for both of the evaluation metrics. 

\subsubsection{Interaction between Topic Structure \& Discourse Relational Structure} We separately parse each of the topic segment of a document through PDTB-styled discourse connective based parser \cite{lin-12} and we collect in-topic keywords. We use a method of collecting all terms on the basis of term-frequency-inverse-document-frequency, then we consider the highly frequent terms using an overlapping 3-turn-windows over the whole segment; we notice patterns that 1. the in-topic keyword density of the topic structure is higher before the commencement of discourse relations and 2. there is a tendency that the explicit connectives do not occur at the beginning of the topic segment. So the fine-grained topics are subjected by the discourse relations. The discourse relations are satisfied if the topic structure constraints are fulfilled. This gives an indication of influence of discourse connectives to link up topic structures together, which helps the global level meaning construction from the local level meanings and also it facilitates disambiguation. Evidences from $Bmr026$ and $Bmr012$ documents are used for this short study. 

 \section{Conclusion}\label{sec:concl}
 
In this study, we propose a laughter information based method of hybrid topic structure for the conversational discourse. There are already several successful approaches for topic segmentation; the proposed approach offers a hierarchy of global and local discourse combining coarse and fine-grained segmentation that hybrids paralinguistic and linguistic information from natural discourses. 

The paper has described a novel training-free algorithm that uses only the paralinguistic information, that is the laughter occurrences to segment the whole conversation into several discourse segments in parallel with two clustering techniques: agglomerative and $K$-medoids. These segmentations are further optimized using a boundary clumping algorithm for the agglomerative case and using an iterative for the $K$-medoids case. Then we choose the best performing clustering based approach for each conversation. Finally through an iterative process we hybridize the fine-level segmentation of a third party Bayesian segmentation tool for each of the coarse level segmentation through clustering. Our proposed approach achieves a hybrid topic structure of coherence that out performs the stand-alone approaches. 

This frame-work can be applicable to the online scenario of spoken language understanding. 
Through the brief study we find that there is an opportunity for an in-depth study of interactions between the topic structure and the other discourse structures, namely discourse relational structure.
\section{Acknowledgement}
This work was primarily done in Idiap Research Institute, Switzerland. The author thanks to Dr. Mathew Magimai Doss and Dr. Afsaneh Asaei for all help.

\bibliographystyle{hacm}
\bibliography{references}

\end{document}